\documentclass[letterpaper, 10 pt, conference]{ieeeconf}  

\IEEEoverridecommandlockouts                              

\usepackage{caption}
\usepackage{algorithm}
\usepackage{algpseudocode}

\usepackage{booktabs} 
\usepackage{graphicx} 
\usepackage{makecell} 
\usepackage{amsmath} 
\usepackage{amssymb} 
\usepackage{xcolor,blindtext} 
\usepackage{subfig}
\usepackage{color} 
\usepackage{balance} 
\usepackage{multirow}

\begin{document}

\title{Iterative PnP and its application in 3D-2D vascular image registration for robot navigation}

\author{Jingwei Song, Keke Yang, Zheng Zhang, Meng Li, Tuoyu Cao, and Maani Ghaffari%
\thanks{J. Song is with United Imaging Research Institute of Intelligent Imaging, Beijing 100144, China
 \texttt{jingwei.song@cri-united-imaging.com}}%
 \thanks{K. Yang, Z. Zhang, M. Li, and T. Cao are with United Imaging, Shanghai, China. \texttt{\{keke.yang, zheng.zhang, meng.li02, tuoyu.cao\}@united-imaging.com}. }
   \thanks{M. Ghaffari is with the University of Michigan, Ann Arbor, MI 48109, USA. \texttt{maanigj@umich.edu}.}%
}
\maketitle
\thispagestyle{empty}
\pagestyle{empty}

\begin{abstract}

This paper reports on a new real-time robot-centered 3D-2D vascular image alignment algorithm, which is robust to outliers and can align nonrigid shapes.
Few works have managed to achieve both real-time and accurate performance for vascular intervention robots. This work bridges high-accuracy 3D-2D registration techniques and computational efficiency requirements in intervention robot applications. We 
categorize centerline-based vascular 3D-2D image registration problems as an iterative Perspective-n-Point (PnP) problem and propose to use the Levenberg-Marquardt solver on the Lie manifold. Then, the recently developed Reproducing Kernel Hilbert Space (RKHS) algorithm is introduced to overcome the \textbf{``big-to-small''} problem in typical robotic scenarios. Finally, an iterative reweighted least squares is applied to solve RKHS-based formulation efficiently. Experiments indicate that the proposed algorithm processes registration over 50 Hz (rigid) and 20 Hz (nonrigid) and obtains competing registration accuracy similar to other works. Results indicate that our Iterative PnP is suitable for future vascular intervention robot applications.
\end{abstract}


\IEEEpeerreviewmaketitle

\section{Introduction}

Endovascular Image-Guided Interventions (EIGIs) is the process of inserting a guidewire/catheter into the femoral artery, which is then threaded under fluoroscopic and angiographic guidance through the vasculature to the site of the pathology to be treated \cite{rudin2008endovascular}. Unlike conventional invasive surgeries, EIGIs play an important role in the modern minimally invasive treatment of heart disease, cancer, and stroke or neurovascular disease, thanks to their lower mortality rate and faster recovery. The minimally invasive EIGIs enables surgeons to deliver drugs, embolic material, or device through the narrow endovascular. In typical EIGIs, rapid-sequential 2D fluoroscopic X-ray imaging is usually heavily relied on for guidance. In addition to the 2D fluoroscopic and angiographic images, pre-operative 3D Digitally Subtracted Angiograms (3D-DSA) or Computed Tomography Angiograms (3D-CTA) can be obtained for surgeons' 3D spatial perception. \par

Although EIGIs are gaining popularity, they cause serious 3D perception problems for surgeons. Surgeons are required to operate in the endovascular with the provided real-time 2D images and without direct 3D visualization. As \cite{mitrovic20183d} points out, surgeons mentally align the interventional tools and blood flow characteristics from real-time 2D images with the displayed 3D vascular morphology. Furthermore, deformation caused by breathing, heart beating, or even patient movement brings more difficulty for surgeons~\cite{10246381}. Thus, extensive research has been conducted in studying automatic alignment algorithms for 3D-2D vascular images by considering 6 Degrees-of-Freedom (DoF) rigid transformation and further nonrigid motion.\par

3D-2D image alignment methods can be categorized as model-driven and data-driven approaches. Model-driven approaches use hand-crafted similarity metrics. \cite{mitrovic20183d}'s comprehensive review categorizes model-driven approaches as intensity-based, feature-based, and gradient-based. These prior-free approaches adopt invariant features in both 3D and 2D images to define the metric, like intensity similarity-based, vascular centerline geometry, image gradient-based, or a hybrid. Data-driven approaches adopt Deep Neural Networks (DNN), which can encode prior information implicitly. Existing data-driven methods~\cite{zheng2018pairwise,liao2017artificial,miao2018dilated} model the entire procedure by predicting the rigid spatial transformation from 3D and 2D pairs in an end-to-end manner. Early works train DNN to predict the transformation in one time~\cite{zheng2017learning} followed by using DNN to incrementally align 3D and 2D shapes~\cite{miao2018dilated,miao2019agent}.\par

Although recent 3D-2D alignment approaches have achieved great progress, their applications in intervention robots remain challenging. First, outliers and incomplete vessel shapes pose great difficulty in alignment. Multi-modal 3D and 2D vascular images are converted to an invariant space for similarity measurement, and the conversion brings outliers. Furthermore, 2D images normally cover a small Field of View (FoV) in reference to the pre-operative 3D images. The missing/redundant features cause partial overlap, matching a big 3D shape to a small 2D shape (\textbf{``big-to-small''}). Existing model-driven approaches with Euclidean losses are especially sensitive to outliers. Secondly, most existing model-based approaches adopt derivative-free optimizers like Powell or Nelder-Mead algorithms. Even though derivative-free optimizers are less sensitive to local minima, they require far more iterations than derivative-based approaches. Recent deep reinforcement learning-based methods~\cite{zheng2018pairwise,miao2018dilated} consume even more computation in their repetitive inference. Lastly, nonrigid vessels bring more obstacles to 3D-2D alignment. Model-driven approaches~\cite{rivest2012nonrigid,chou2012real,zhang2007patient} formulate the alignment with much larger nonrigid state space and require heavy computation in the minimization process. Existing nonrigid data-driven approaches~\cite{guan2020transfer,guan2019deformable,nakao2022image} are not efficient in real-time implementation.

In this paper, we aim to achieve the first robot-centered 3D-2D vascular image alignment algorithm, which runs in real-time, is robust to outliers, and can align shapes in nonrigid manner. We first categorize the 3D-2D aligning problem as an Iterative Perspective-n-Point (PnP)~\cite{haralick1994review,Lepetit_160138}, then adopt 2D Reproducing Kernel Hilbert Space (RKHS) loss to handle outliers. Considering that a typical second-order derivative solver cannot optimize RKHS loss, an Iterative Reweighted Least Squares (IRLS) combined with Levenberg–Marquardt is applied in solving RKHS formulation. It substitutes conventional derivative-free optimizers with a second-order optimizer, leading to a reduced number of iterations. Moreover, experiments indicate that RKHS loss is more robust to outliers than Euclidean loss, especially in the \textbf{``big-to-small''} case. 

In particular, the contributions of this work are as follows.

\begin{enumerate}
    \item 3D-2D vascular image alignment is formulated as an Iterative PnP with RKHS loss. The formulation shows increased robustness to outliers compared with conventional Euclidean loss.
    \item This is the first work adopting an exact second-order derivative-based optimizer for real-time alignment. IRLS is applied to minimize RKHS loss as the second-order optimizer.
    \item Moreover, this work combined the nonrigid and RKHS formulations into a joint framework.
\end{enumerate}

\section{Literature review}

3D-2D alignment algorithms can be classified as model-driven and data-driven. Both are based on the assumption that the projected 3D image should be the most similar to the target 2D image with the correct intrinsic projection matrix, pose, and optional nonrigid parameter.\par

Model-driven approaches formulate the problem by estimating the sensor pose and optional nonrigid parameter with the extracted handcrafted invariant features. According to~\cite{mitrovic20183d}, model-driven approaches consist of intensity-based, feature-based, and gradient-based approaches. Intensity-based algorithms project the 3D image with Digitally Reconstructed Radiographs (DRRs)~\cite{hipwell2003intensity} or Maximum Intensity Projection (MIP)~\cite{kerrien1999fully} and measure the image similarity based on metrics like normalized cross-correlation. Intensity-based algorithms require heavy computation in the iterative 3D to 2D projection process in the full image domain~\cite{mitrovic20183d}. Feature-based approaches extract invariant semantic features like centerline, point orientation, and bifurcation. \cite{aylward2003registration} uses the Gaussian kernel function and estimates the 6 DoF transformation by maximizing the sum of kernels in 2D. \cite{groher2009deformable} extends the rigid matching to nonrigid matching and defines the loss as a 2D Euclidean sum. \cite{rivest2012nonrigid} minimizes the projection error of 2D vessel tree skeleton in a two-step process (rigid and nonrigid). Gradient-based methods build the metric by exploiting the fact that the 2D gradient and 3D projected gradient are perpendicular on the aligned vessel. Experiments in~\cite{mitrovic20183d} validate that intensity-based methods are the most accurate, while others are with similar accuracy. Three algorithms~\cite{rivest2012nonrigid,mitrovic20133d,jomier20063d} achieve around $0.5$s while the rest requires over $10$s. \cite{groher2009deformable} adopted the derivative-based algorithm Broyden–Fletcher–Goldfarb–Shanno (BFGS) as the solver. \cite{mitrovic20133d} used a derivative-free algorithm (Powell) but implemented on GPU. Although not reported, \cite{jomier20063d} is expected to be slow because it is based on a Genetic Algorithm. \par

Data-driven algorithms learn the similarity metric from the training data and predict the optimal transformation in an end-to-end manner. The network is trained on a simulated predicted 2D data set from the pre-operative 3D data set and applied to the real data set. Early works~\cite{zheng2018pairwise,zheng2017learning} used multi-range networks for coarse-to-fine sensor pose estimation. Each network covers a certain range of the pose. \cite{guan2019deformable} proposes a multi-channel convolution DNN that integrates multiple phases caused by the breathing and heartbeat of patients. Its inference consumes 4 milliseconds (ms), thanks to the small size of the neural network. The following studies used reinforcement learning to align shapes iteratively. A simplified Q-learning, based on the Markov decision process, was applied in a step-by-step manner~\cite{miao2018dilated,miao2019agent}. \cite{guan2020transfer} adopted transfer learning to bridge the domain gap between training and testing data sets.  Although DNN-based methods have been reported to be accurate, most of them require long training and inference time, and some are difficult to reproduce. \par

According to our knowledge, only \cite{groher2009deformable} adopted a derivative-based BFGS solver in optimal pose searching procedure. Derivative-based searching, especially second-order derivative algorithm, guarantees real-time searching in a fixed number of iterations. This research follows~\cite{rivest2012nonrigid}'s roadmap by aligning 3D-2D images' vascular centerlines and adapts it to EIGIs.\par

\section{Methodology}
This work first conducts 3D and 2D blood vessel centerline segmentation and then estimates the optimal rigid and nonrigid transformation for 3D and 2D vessel alignment. We will reveal that 3D-2D centerline matching is a special case of a well-researched topic, that is, PnP, in the computer vision community. Several state-of-the-art strategies are coupled to realize a real-time robust 3D to 2D vessel alignment. The proposed workflow ensures robust and prior-free methods for intervention robots.\par

\subsection{Vessels segmentation}

This research chose vascular-based data alignment for intervention robot applications for two reasons. First, blood vessel centerlines are invariant and observable tissues in both 3D and 2D images. The centerline matching procedure is always differentiable, while intensity-based and gradient-based alignments are not always differentiable if the initial alignment is too far away. Second, 3D and 2D vessel centerline segmentation algorithms are heavily studied and are robust for robotic applications. \par

Widely used Otsu's method~\cite{otsu1979threshold} was applied for fast 2D vessel segmentation. DNN-based method from commercial software provided by United Imaging of Health Co., Ltd is used for 3D vessel segmentation. Other prior-based DNNs~\cite{li2022dual,moccia2018blood} are also applicable for 3D/2D image segmentation.\par

\subsection{Iterative PnP formulation}

After vessel segmentation, the 3D-2D registration algorithm can be summarized as an approximate Iterative PnP. PnP is the algorithm for estimating the optimal 6 DoF sensor pose of a set of 3D points given the calibrated projection matrix and 3D-2D correspondences that align projected 3D points and 2D points~\cite{fischler1981random}. To realize the algorithm, correspondences should be established based on the closest projections search given the initial sensor pose. Thus, iterative PnP differs from authentic PnP, which associates points based on feature points similarity. Data association process in the iterative PnP is similar to the Iterative Closest Point (ICP) algorithm. Another difference from authentic PnP is that vessels may deform in surgeries like percutaneous coronary intervention (mainly from heartbeat) and transarterial chemoembolization (mainly from breath). Therefore, nonrigid parameterization is an option in addition to rigid pose estimation.\par

PnP is a fundamental tool~\cite{haralick1994review,Lepetit_160138} for searching the optimal camera (sensor) pose $\mathbf{T}^* \in \mathrm{SE}(3)$ with know 3D point $\mathbf{p}_i \in \mathbb{R}^4$ ($\Omega$ is the set of source point index) and 2D $\mathbf{q}_i \in \mathbb{R}^3$ (both $\mathbf{p}_i$ and $\mathbf{q}_i$ are in homogeneous coordinate) alignment. Thus, the correspondence $\mathbf{p}_i$ and $\mathbf{q}_i$ are aligned point pair. Authentic PnP is formulated as:\par
\begin{equation}
	\label{Eq_pnp}	\mathbf{T}^*=\operatorname{argmin}_{\mathbf{T}} \sum_{i \in \Omega}||\pi(\mathbf{T}\mathbf{p}_i,\mathbf{K})-\mathbf{q}_i||^2_2,
\end{equation}

\noindent where $\mathbf{K} \in \mathbb{R}^{3\times4}$ is the camera intrinsic matrix and $\pi(\mathbf{T}\mathbf{p}_i,\mathbf{K})$ is the 3D to 2D pin-hole camera projection function defined as:

\begin{equation}
\pi(\mathbf{T}\mathbf{p}_i,\mathbf{K}) = \mathbf{K}\mathbf{T}\mathbf{p}_i.
\end{equation}
 
Similar to \eqref{Eq_pnp}, \cite{rivest2012nonrigid}~defines 2D points' Euclidean distance sum as the loss function and obtain optimal sensor pose as
\begin{equation}
	\label{2Ddis_object_func_0_0}	\mathbf{T}^*=\operatorname{argmin}_{\mathbf{T}} \sum_{i \in \Omega}||\pi(\mathbf{T}\mathbf{T}_{dsa\_cb}(\mathbf{p}_i-\mathbf{c}),\mathbf{K})-\mathbf{q}_j||^2_2,
\end{equation}

\noindent where $\mathbf{T}_{dsa\_cb} \in \mathrm{SE}(3)$ is the pre-calibrated transformation matrix from pre-operative coordinate (CT for example) to intra-operative coordinate (DSA for example), $\mathbf{p}_i$ is the 3D source point in homogeneous coordinate and $\mathbf{q}_j$ is the corresponding 2D target point in homogeneous coordinate by closest distance searching. Instead of known correspondences, the temporal correspondences between groups $\{\mathbf{p}_i\}$ and $\{\mathbf{q}_j\}$ are determined by point-to-point closest Euclidean distance searching similar to ICP. \cite{rivest2012nonrigid} relaxed the rotation in $\mathbf{T}$ as an affine matrix for modelling simple deformation.\par

We define that \eqref{Eq_pnp} with unknown $\mathbf{p}_i$ and $\mathbf{q}_i$ pairs (or $\mathbf{q}_j$ to be specific) can be defined as \textbf{Iterative PnP}. Robot and computer vision communities seldom handle this scenario because the correspondences can be obtained with well-known image corner points matching algorithms. It only happens in cases like vascular image registration. In contrast to PnP, Iterative PnP brings two issues: more outliers and more local minima. The following RKHS strategy handles outliers well. For local minima, our experiments show that obtaining the correct pose is difficult, but the registration task remains unaffected.

\begin{figure}[t]
    \centering
    \includegraphics[width=0.9\columnwidth]{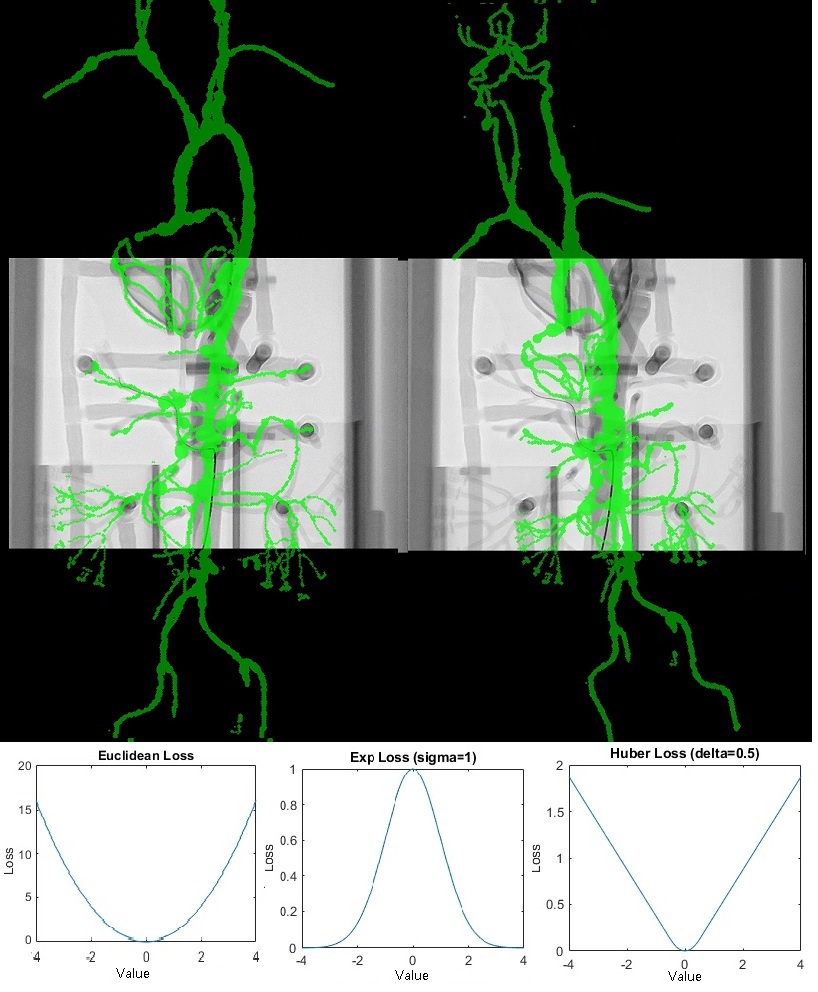}
    \caption{The upper two figures show the typical \textbf{``big-to-small''} alignment. The green shape is the projection of the pre-operative 3D vessel, which is in large size. The upper left figure is the proposed Iterative PnP on RKHS space, which is robust to \textbf{``big-to-small''} alignment. The upper right figure is our implementation of ~\cite{rivest2012nonrigid}, which uses the Huber loss. The bottom figures are three loss functions. Only the ``Exp'' (stands for exponential) loss manages to lower the impacts of ``outlier'' points.}
    \label{fig_big_to_small}
    \vspace{-4mm}
\end{figure}

\subsection{Outlier rejection and RKHS}

Equation~\eqref{2Ddis_object_func_0_0} is the widely used L2 norm loss for authentic PnP and ICP problems and works well in many situations. Although it can be solved with Gauss-Newton or quasi-Newton solver, it suffers from a serious issue widely known as \textit{partially overlap}. Specifically, the pre-operative 3D blood vessel is much larger than the intra-operative 2D blood vessel, as \textbf{``big-to-small''} in this article. A massive number of unpaired source points $\mathbf{p}_i$ are wrongly matched to target points $\mathbf{q}_i$. Since the wrong pairs have large distances, the sum of errors is large and brings wrong derivative weighting in searching for optimal $\mathbf{T}^*$. Although M-estimators~\cite{maronna1976robust} reduce the influence of large distance pairs, the matching is still heavily influenced by outliers. Conventional 3D-3D ICP can circumvent this issue by reversing source and target point clouds (that is ``small-to-big''). In 3D-2D matching, however, reversing is impossible because the key accelerating ingredient, the K-D tree-based correspondence searching, should only be built on the static 2D vessel. As Fig. \ref{fig_big_to_small} shows, many points outside of the FoV are outliers and Euclidean loss in \eqref{2Ddis_object_func_0_0} suffers from large outliers (upper right figure) and then wrong derivatives. Huber loss only manages to reduce the Binomial distance-to-loss mapping to linear mapping. \par

Gaussian kernels or Gaussian Mixture Model (GMM)~\cite{biber2003normal,magnusson2007scan,jian2010robust} overcomes the partial overlapping problem (big-to-small in 3D-2D vessel registration case) based on its special distance-to-loss mapping manner. \cite{aylward2003registration} is the first work applying 3D-2D vessel registration based on Gaussian kernel. This work uses a Gaussian kernel to measure alignment in 2D space and demonstrates its efficacy. During our tests, we also noticed that Gaussian kernel~\cite{aylward2003registration} brings better accuracy and robustness to 3D-2D blood vessel matching. Surprisingly, most centerline-based studies ignore it and use Euclidean loss instead. One possible explanation is that Gaussian kernel cannot be transformed into authentic least-square form, and real-time applications like \cite{rivest2012nonrigid} cannot convert Gaussian kernel loss to BFGS solver.\par

Continuous Visual Odometry (CVO)~\cite{clark2021nonparametric} is the first work applying Reproducing Kernel Hilbert Space (RKHS) theory in an ICP problem. Different from \cite{biber2003normal} using discrete Gaussian kernel space, CVO~\cite{clark2021nonparametric} enforces continuous GMM field and adopts 4th order Taylor expansion to retrieve an optimum step for the first-order derivative-based gradient descend. It achieves 2-5 Hz on modern CPU with parallel computation and for typical LiDAR-based point cloud registration. \cite{zhang2021new} further reduces CVO's time consumption with GPU acceleration. Following~\cite{zhang2021new}, this article modifies RKHS~\cite{clark2021nonparametric} and obtain optimal parameters $\mathbf{T}^*,\theta^*$ by jointly optimizing\par
\begin{equation}
\begin{aligned}
	\label{GMM_object_func}
	&
 \operatorname{max}_ {\mathbf{T},\theta}  E_{data} \quad \text{and} \quad \operatorname{min}_ {\mathbf{T}, \theta} E_{init} + \lambda E_{reg}, \text{ such that}\\
 &E_{data}= \\
 &\sum_{i \in \Omega}\mathrm{w}_i\!\sum_{j \in \Delta(i)}\exp \left(-\frac{\lVert \pi(\mathbf{T}\Gamma(\mathbf{T}_{dsa\_cb}\mathbf{p}_i-\mathbf{c},\theta),\mathbf{K})-\mathbf{q}_j \rVert^2_2}{2\ell^2}\right)\\
  &E_{init}= \lVert \operatorname{log}(\mathbf{T}^{-1}_0\mathbf{T})^{\vee} \rVert_2^{2}\\
  &E_{reg} = \mathrm{w}_1 \sum_{i \in \Omega} \left\lVert \mathbf{r}_i \right\lVert^2_{2}  +  \mathrm{w}_2 \sum_{i \in \Omega}\sum_{j\in \{i-1,i+1\}} \left\lVert \mathbf{r}_i - \mathbf{r}_j \right\lVert^2_{2} \\
    & \qquad + \mathrm{w}_3 \sum_{i \in \Omega}\sum_{j \in \Omega_i} \left\lVert \mathbf{r}_i - \mathbf{r}_j \right\lVert^2_{2} ,
\end{aligned}
\end{equation}


\noindent where $\mathrm{w}_i$ is the weight and uniformly set as 1 in this research, $\Delta(i)$ is the set of correspondences to $\mathbf{p}_i$, $\mathbf{T}_0$ is the initial pose or pose in previous state and $\ell$, $\lambda$, $\mathrm{w}_1$, $\mathrm{w}_2$, $\mathrm{w}_3$ are hyperparameters, and $\theta$ is the deformation parameter. $\Omega_i$ is neighboring points' index set for $\mathbf{p}_i$. $(\cdot)^\vee$ convert lie algebra $\mathfrak{se}(3)$ to 6-vector $\mathbb{R}^6$ and defines the distance on $\mathrm{SE}(3)$ manifold. It should be pointed out that $E_{init}$ is adopted because $\mathbf{T}^*$'s z-direction increases enormously, and the 3D shape's projection shrinks to a point and matches to an arbitrary target point. This phenomenon represents one scenario of outlier and thus should be handled with $E_{init}$. $\ell$ is defined as the maximum $||\pi(\mathbf{T}\Gamma(\mathbf{T}_{dsa\_cb}\mathbf{p}_i-\mathbf{c},\theta))-\mathbf{q}_j,\mathbf{K}||^2_2$ and shrinked by half every 5 iterations.\par 

In rigid cases, the nonrigid function is defined as \mbox{$\Gamma(\cdot,\theta) = \cdot$}. Meanwhile, $\lambda_2$ should be set to zero. For nonrigid Iterative PnP, we follow \cite{rivest2012nonrigid} by realizing $\Gamma(\mathbf{p}_i,\theta)$ as:

\begin{equation}
\begin{aligned}
	\label{2Ddis_object_func_Gamma}
    &\Gamma(\mathbf{p}_i,\theta) = \mathbf{p}_i + \mathbf{r}_i,  \text {\  such that } \mathbf{r}_i \in \theta,
\end{aligned}
\end{equation}

\noindent where $\theta=\{\mathbf{r}_i|i\in\Omega\}$ and $\mathbf{r}_i \in \mathbb{R}^4$.
\eqref{GMM_object_func} differs from \cite{aylward2003registration} as the nonrigid deformation field, 2D RKHS space supported by $\Delta_{i}$, initial sensor pose $\mathbf{T}_0$ regularization and approximate second-order derivative fastening covered in next section.

\subsection{Approximate second-order optimizer for RKHS space}

Real-time implementation is the key requirement for intervention robot applications. Most studies focus on achieving high accuracy instead of computational efficiency and thus adopt derivative-free easy-to-use solvers like the Powell or Nelder-Mead method. Only \cite{groher2009deformable} adopts a second-derivative-based solver BFGS for fast optimization. This work attempts to solve the original multi-objective function \eqref{GMM_object_func} by vectorizing the original problem in an reweighted least squared manner. Specifically, considering that Gaussian kernel cannot be written in binomial form and thus least squares, \cite{clark2021nonparametric} uses first-order gradient descent and retrieve the optimal step by approximating Gaussian kernel as 4th order Taylor expansion. However, their experiment indicates that their accelerated first-order gradient descent still requires a massive amount of iteration. This work follows the latest progress~\cite{clark2021nonparametric} and adopts the IRLS method in optimization. In the iterative optimization of \eqref{GMM_object_func}, the searching direction and step size of ${\mathbf{T}^{n}, \theta^{n}}$ in step $n$ is determined 
\begin{equation}
\begin{aligned}
	\label{GMM_object_func_irls}
	&\mathbf{T}^{n+1},\theta^{n+1}= \operatorname{min}_ {\mathbf{T},\theta} \sum_{i \in \Omega}\mathrm{w}_i \sum_{j \in \Delta(i)}\mathrm{w}_{ij}^{n} \lVert\pi(\mathbf{T}^{n}\\
&\cdot(\Gamma(\mathbf{T}_{dsa\_cb}\mathbf{p}_i - \mathbf{c},\theta^{n}))-\mathbf{q}_j\rVert^2_2 + \lambda_1 \mathrm{E}_{init} + \lambda_2 \mathrm{E}_{reg},
\end{aligned}
\end{equation}

\noindent where $\lambda_1$ and $\lambda_2$ are the hyperparameters and $\mathrm{w}_{ij}^{n} = \exp \left(-\frac{\lVert \pi(\mathbf{T}^{n}\Gamma(\mathbf{T}_{dsa\_cb}\mathbf{p}_i-\mathbf{c},\theta^{n}),\mathbf{K})-\mathbf{q}_j \rVert^2_2}{2\ell^2}\right)$ and fixed as the weight. $\mathbf{T}^{n}$ and $\theta^{n}$ are the state estimated in step $n$ and indifferentiable in $\mathrm{w}_{ij}^{n}$. The state $\mathbf{T}$ and $\theta$ in \eqref{GMM_object_func_irls} are initialized with $\mathbf{T}^{n}$ and $\theta^{n}$. \textbf{It should be noted that although the searching direction and step are determined by local descending of \eqref{GMM_object_func_irls} in each step, the main objective is to maximize the sum of RKHS.} In practice, the direction and size in \eqref{GMM_object_func_irls} is searched in least square form for . \par 

\subsection{Technical details}

Following \cite{rivest2012nonrigid}, 6 DoF rigid pose $\mathbf{T}^*$ in \eqref{GMM_object_func} is first estimated, and nonrigid 
field $\theta^*$ is then estimated for refinement. Moreover, rigid transformation $\mathbf{T}$ is incrementally optimized on $\mathrm{SE}(3)$ manifold~\cite{boumal2023introduction} in contrast to Euler angles and $\mathbb{R}^3$ translation in existing reseaches. Optimization on continuous $\mathrm{SE}(3)$ manifold avoids gimbal lock degeneration on Euler angles space by following geodesics along the manifold for search. Levenberg-Marquardt is adopted as the minimizer. \par

\section{Experiments}

\subsection{Experimental Setup}

The proposed algorithm was tested on a simulation data set, an \textit{ex-vivo} data set, and an \textit{in-vivo} data set\footnote{We strongly recommend readers to watch the attached video for the complete \textit{in-vivo} and \textit{ex-vivo} experiments.}. The simulation data set was generated by enforcing a simulated $\mathrm{SE}(3)$ transformation on segmented 3D vessel centerlines. The chosen algorithm aligned it with its original projection without the simulated $\mathrm{SE}(3)$ transformation. As~\cite{rivest2012nonrigid} points out, testing on simulations characterizes the performance of the various algorithms in an ideal
scenario with known sensor pose and without noises. The adopted \textit{in-vivo} data sets of 6 patients are P1 (44), P2 (60), P3 (39), P4 (53), P5 (45), P6 (37), where the number in bracket records the size of images. An \textit{ex-vivo} data set was also generated by scanning a phantom using United Imaging's uAngio 960 devices. 3D pre-operative vessel centerlines were obtained by segmenting a pre-operative CTA image. Then, live images were collected for real-time alignment. \par

The proposed algorithm was compared with widely accepted prior-free approaches AutoMask~\cite{steininger2012auto}, DT-ICP~\cite{rivest2012nonrigid}, RGRB~\cite{markelj2008robust} and Normalized ICP (NICP)~\cite{aylward2003registration}. The original DT-ICP~\cite{rivest2012nonrigid} uses BFGS as the second-order solver, while our implementation adopts the Levenberg-Marquardt solver on the Lie manifold for better convergence and faster speed. Our experiments show faster speed and less iteration than BFGS. Pair-wise DNN~\cite{zheng2018pairwise} is implemented for comparison. For DNN-based approaches, we admit that reinforcement learning-related approaches~\cite{miao2018dilated,miao2019agent} are extremely difficult to implement and train. We fail to realize these approaches. As~\cite{meng2022weakly} pointed out, these complicated DNN-based approaches suffer from ``hardness of obtaining the ground-truth transformation parameters'' and ``real intra-operative DSAs
can be incomplete''. \cite{meng2022weakly} shifted the research direction back to two-step-based image-to-pose regression. \par 

A commercial laptop ALIENWARE M17 R4 (i7-10870H and 32Gb RAM) is used. Its GPU, GeForce RTX 3060 (6Gb), is used for Pair-wise DNN only. As no open-source code is available, all these approaches were implemented by ourselves. AutoMask was implemented on python3 because its derivative-free optimizer, Nelder-Mead, is slow and impossible for real-time performance. RGBR was implemented on Matlab and its time consumption is meaningless. DT-ICP, NICP, and Iterative PnP were implemented in C++ and wrapped as a package in Robot Operating System (ROS)~\cite{quigley2009ros}. During all experiments, both 3D and 2D vessel sizes are within 1500 - 3000 points. The hyperparameters $\lambda_1$, $\lambda_2$, ${w}_1$, ${w}_2$ and ${w}_3$ are set as $100$, $1$, $0.1$, $10$ and $1$.\par

Following previous research, all the algorithms were compared based on Projection Residual (PR), Gross Failure Rate (GFR), and time consumption. PR measures the Root Mean Square Error of all the projected source points. GFR accounts for test cases with PR greater than the selected threshold. Median, $75\%$th percentile, and $95\%$th percentile PRs are reported. In quantitative experiments, all points were selected as anchor points. Since the aim is limited to centerline registration as other works do, Otsu-based extraction's time consumption which is around $400$ ms for image size $512 \times 512$, is not counted. \par
 
\subsection{Accuracy and Robustness}

\begin{figure}[t]
    \centering
\includegraphics[width=0.6\columnwidth]{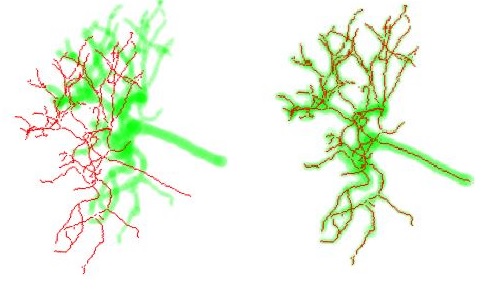}
    \caption{Presented is a sample synthetic data set. The red line is the synthetic 2D vascular centerline. Green shape is the transformed and projected 3D shape. The left shape is the projection with a fixed initial pose, and the right shape is projected with a fixed pose.}
    \label{fig_synth_results}
    \vspace{-4mm}
\end{figure}

The simulation experiment was conducted by sampling rigid pose 100 times with angular standard deviation $2^\circ$ and translation standard deviation $5mm$. Simulation experiments guarantee one-to-one correspondence without noise. Fig. \ref{fig_synth_results} shows one sample Iterative ICP result of the simulated 2D centerline (red) and 3D projection based on the selected pose. Table \ref{Table_simu_result} shows that DT-ICP performs similarly as Iterative PnP (rigid) and Iterative PnP in the simulation data sets.\par 

\begin{figure*}[!h]
		\centering
		\subfloat{
			\begin{minipage}[]{1\textwidth}
				\centering
				\includegraphics[width=1\linewidth]{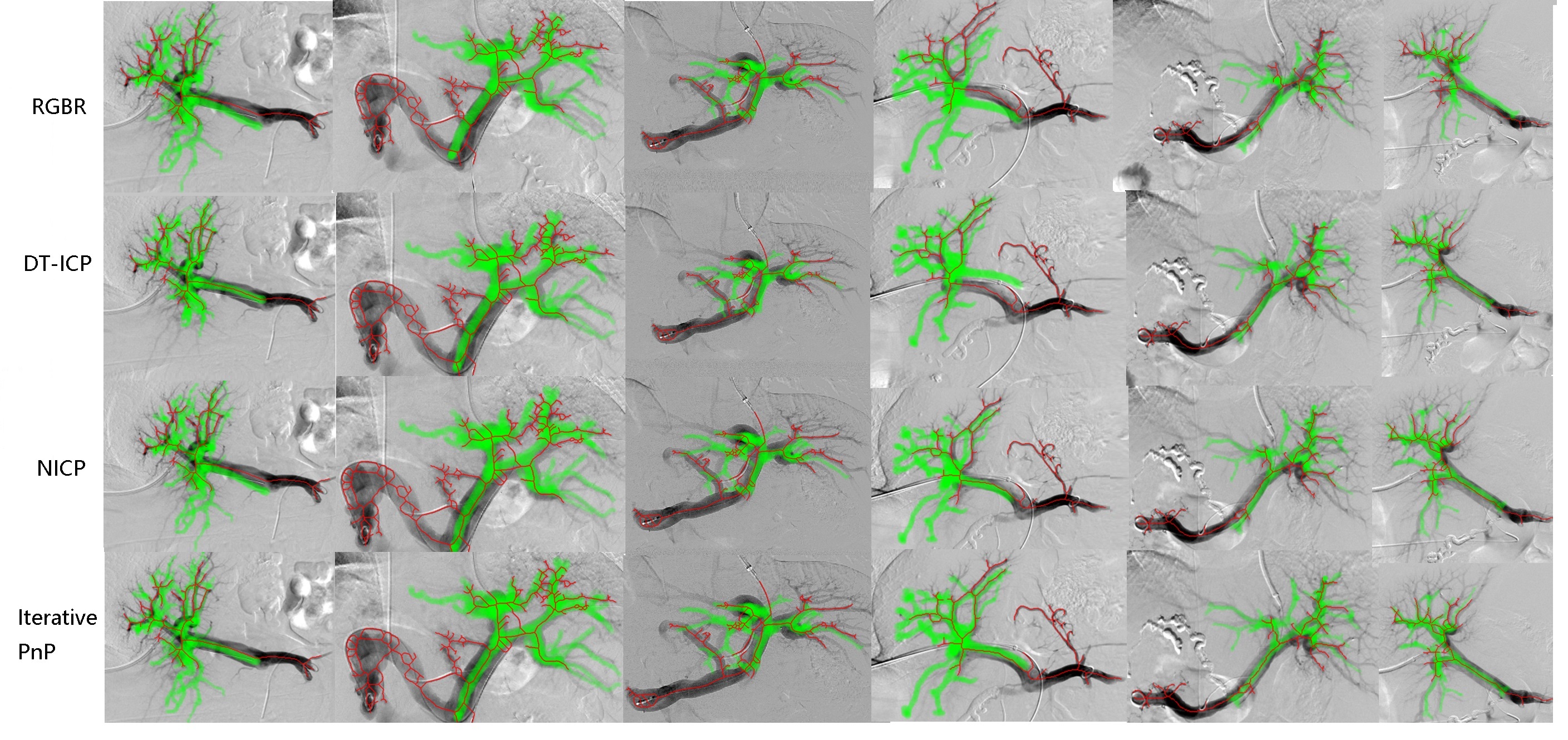}
			\end{minipage}
		}
		\caption{Illustrated is the alignment comparisons of RGBR, DT-ICP, NICP and Iterative PnP. Green shape is the projection of the pre-operative 3D vessel and red shape is the 2D centerline of the vessel. The 6 columns represents the sample results of the 6 patients.}
		\label{fig_invivo_results}
  \vspace{-3mm}
	\end{figure*}

\begin{figure}[!h]
		\centering
		\subfloat{
			\begin{minipage}[]{0.5\textwidth}
				\centering
				\includegraphics[width=0.9\linewidth]{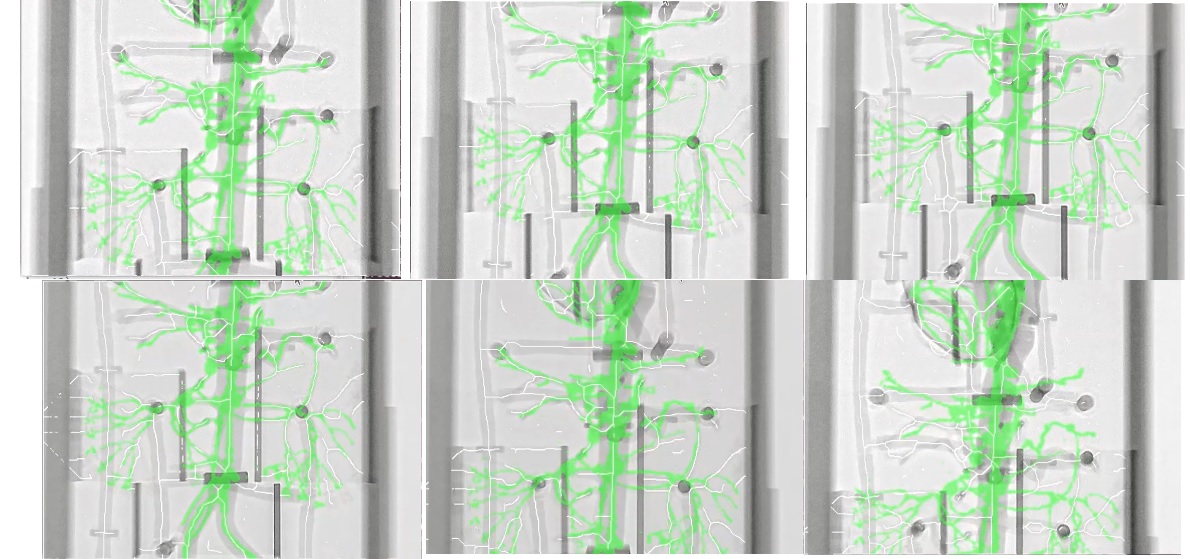}
			\end{minipage}
		}
		\caption{The figure shows one sample of the \textit{ex-vivo} experiments. Sequential registered 3D-2D phantom shapes masked on X-ray images are demonstrated. The green shapes are the projected 3D vessels. Other \textit{ex-vivo} experiments can be found in the attached video.}
		\label{fig_exvivo_results}
  \vspace{-4mm}
	\end{figure}

\textit{In-vivo} and \textit{ex-vivo} experiments were implemented. Fig. \ref{fig_invivo_results} visualizes sample comparisons of the registration of the 6 \textit{in-vivo} data sets. Fig. \ref{fig_big_to_small} demonstrates that the proposed RKHS loss handles big-to-small issues well, while DT-ICP fails in the alignment. Fig. \ref{fig_exvivo_results} shows a sample sequential registration in X-ray-based \textit{ex-vivo} experiments. Visually, the proposed Iterative PnP achieves slightly better registration than competing algorithms. All qualitative results validate that Iterative PnP achieves slightly or even better results than competing methods.\par

\begin{table}[]
\centering
\caption{The table reports the average pose of the simulation data by enforcing noise with angular standard deviation $2^\circ$ and translation standard deviation $5mm$. GFR refers meanPR greater than $5$ pixels.}
\begin{tabular}{llllll}
\toprule
\multicolumn{1}{c}{\multirow{2}{*}{Method}} & \multicolumn{1}{c}{\multirow{2}{*}{GFR}} & \multicolumn{1}{c}{\multirow{2}{*}{\begin{tabular}[c]{@{}c@{}}MeanPR\\ (mm)\end{tabular}}} & \multicolumn{2}{c}{Percentile (mm)} & \multicolumn{1}{c}{\multirow{2}{*}{\begin{tabular}[c]{@{}c@{}}Run time\\ (ms)\end{tabular}}} \\ \cline{4-5}
\multicolumn{1}{c}{}                        & \multicolumn{1}{c}{}                     & \multicolumn{1}{c}{}                                                                     & 95\%             & 75\%             & \multicolumn{1}{c}{}                                                                         \\ 
\midrule
AutoMask                                & $0\%$                     &  0.85                                                                                         &   1.50               &    1.06              & 97004.0                                                                                           \\
Pair-wise DNN                                & $0\%$                     &  1.74                                                                                         &   2.54               &    2.02              & 311.3                                                                                            \\
RGBR                                        & \multicolumn{1}{r}{$22\%$}                     & \multicolumn{1}{l}{2.21}                                                                     &      3.92       &    1.71              & -                                                                                            \\
DT-ICP (rigid)                                      & $54\%$                 & 1.80                                                                                   & 6.65           & 3.49           & 13.2                                                                                         \\
DT-ICP                                  & $0\%$                                     & 0.42                                                                                    & 0.72            & 0.42            & 39.4                                                                                         \\
NICP                                        & $8\%$                                     & 0.50                                                                                   & 3.11           & 1.14           & 385.0                                                                                        \\
Iterative PnP (rigid)                               & $4\%$                                     & 0.39                                                                                   & 0.84           & 0.56           & 16.9                                                                                         \\
Iterative PnP                           & $0\%$                                     & 0.37                                                                                    & 0.69            & 0.44            & 45.5                                                                                         \\ 
\bottomrule
\end{tabular}
\label{Table_simu_result}
\vspace{-4mm}
\end{table}

\begin{table}[]
\centering
\caption{The table shows the accuracy of the \textit{in-vivo} data sets. Median PR is the median of PR of all source points. GFR refers meanPR greater than $5mm$. }
\begin{tabular}{lrllll}
\toprule
\multicolumn{1}{c}{\multirow{2}{*}{Method}} & \multicolumn{1}{c}{\multirow{2}{*}{GFR}} & \multicolumn{1}{c}{\multirow{2}{*}{\begin{tabular}[c]{@{}c@{}}Median PR\\ (mm)\end{tabular}}} & \multicolumn{2}{c}{Percentile (mm)} & \multicolumn{1}{c}{\multirow{2}{*}{\begin{tabular}[c]{@{}c@{}}Run time\\ (ms)\end{tabular}}} \\ \cline{4-5}
\multicolumn{1}{c}{}                        & \multicolumn{1}{c}{}                     & \multicolumn{1}{c}{}                                                                       & 95\%             & 75\%             & \multicolumn{1}{c}{}                                                                         \\ 
\midrule
AutoMask                              &   $100\%$                                         &  \multicolumn{1}{c}{17.01}                                                                                         &        29.75           &      23.26            & 150302.2                                                                                            \\
Pair-wise DNN                              &     100\%                                       &  \multicolumn{1}{c}{153.25}                                                                                         &      437.24             &          168.60        &       529.0                                                                                      \\
RGRB                                        &      $100\%$                                    & \multicolumn{1}{c}{5.99}                                                                   & 38.85            & 12.16            & -                                                                                            \\
DT-ICP                                      & \multicolumn{1}{l}{$16\%$}                                     & \multicolumn{1}{c}{4.68}                                                                                       & 31.22            & 9.613            & 13.2                                                                                         \\
NICP                                        & \multicolumn{1}{l}{$56\%$}                 & \multicolumn{1}{c}{3.49}                                                                                       & 37.09            & 10.52            & 385.1                                                                                        \\
Iterative PnP                               & \multicolumn{1}{l}{$11\%$}                 & \multicolumn{1}{c}{3.74}                                                                                       & 29.92            & 10.34            & 16.9                                                                                         \\ \bottomrule
\end{tabular}
\label{Table_invivo_result}
\end{table}

\subsection{Time consumption}
Table \ref{Table_simu_result} and Table \ref{Table_invivo_result} show that both DT-ICP and Iterative PnP manage over 50 Hz (rigid) and 20 Hz (nonrigid). The original DT-ICP~\cite{rivest2012nonrigid} reports around 2 Hz. Our implementation with Levenberg–Marquardt solver on Lie manifold requires much less iteration than their BFGS. Other reasons include computational devices and data set differences. Although the proposed Iterative PnP is around $25\%$ slower than DT-ICP, it achieves higher accuracy and better robustness.\par

\subsection{Local minima in Pose Estimation}

\label{sec_local_min}

It is interesting to reveal that Iterative PnP suffers from heavy local minima, making the estimated pose incorrect in most experiments.
Table \ref{Table_local_minima} shows the relationship between the sensor pose disturbance level and pose estimation accuracy of our proposed algorithm. It indicates that robot pose can be estimated correctly with fewer local minima if the initial pose is good (small pose disturbance in Table \ref{Table_local_minima}). Table \ref{Table_local_minima} also suggests that point-wise registration is still successful even though the estimated pose is wrong. A test on 3D-3D (with depth of target data provided) registration of RKHS formulation \cite{clark2021nonparametric} shows that RKHS accurately estimates pose. Therefore, we hypothesize that Iterative PnP gains an extra amount of local minima and should be credited to the unobservable depth for 2D vascular centerlines compared with the notorious nonconvex ICP process. This conclusion should be further validated.

\begin{table}[]
\centering
\caption{The table reports the average pose error of the simulation data with different disturbances, which is in different levels of Gaussian noises (``Std'' refers to standard deviation). MeanPR and GFR shows that registration is still satisfying. Pose errors indicate that the algorithm fails. 1 pixel is equivalent to 0.30 mm. ``Ang Std'' refers angular standard deviation (in $^\circ$). ``Trans Std'' refers translation standard deviation (in $mm$). GFR refers MeanPR greater than $5$ pixels.}
\begin{tabular}{lllll}
\toprule
Ang Std & Trans Std & MeanPR (mm) & GFR  & Average Pose Error \\
\midrule
0.50              & 1.00                    & 0.10         & $0.0\%$                    & $0.10^\circ$, $1.81 mm$  \\
1.00              & 3.00                    & 0.28         & $0.0\%$                    & $0.29^\circ$, $5.31 mm$   \\
2.00              & 5.00                    & 0.43         & $0.0\%$                    & $0.46^\circ$, $7.09 mm$   \\
3.00              & 8.00                    & 0.82         & $0.0\%$                    & $0.68^\circ$, $12.48 mm$  \\
5.00              & 10.00                   & 1.51         & $21.0\%$                    & $3.40^\circ$, $14.98 mm$ \\
\bottomrule
\end{tabular}
\label{Table_local_minima}
\vspace{-4mm}
\end{table}

\subsection{Limitations \& Future Works}
The major issue of Iterative PnP is that its accuracy and robustness are heavily affected by the objective function's nonconvexity. Section \ref{sec_local_min} reveals that Iterative PnP is not suitable for sensor pose estimation. Future works can be devoted to overcoming local minima with methods like Branch-and-Bound strategy~\cite{olsson2008branch} or gradual relaxation~\cite{yang2020graduated}. Better convergence not only estimates 6 DoF pose correctly but also provides better registration (median PR shown in Table \ref{Table_local_minima}). However, the shape alignment can still yield accurate 3D-2D shape alignment.

\section{Conclusion}

This research proposes an Iterative PnP algorithm for robot navigation tasks in EIGIs. Results show that the proposed algorithm achieves real-time alignment over 20 Hz (nonrigid) or 50 Hz (rigid) based on modern laptops. The proposed algorithm achieves similar accuracy and is robust to outliers compared with existing works, especially in typical \textbf{``big-to-small''} vascular centerline registration scenarios. Considering that our Iterative PnP is more robust, slightly more accurate, and computationally efficient, our proposed method is suitable for future EIGIs and intervention robot applications. This work is in the process of commercialization by the company United Imaging of Health Co., Ltd. \par


\balance
{\small
		\bibliographystyle{ieeetr}
		\bibliography{bib/strings-abrv,bib/ieee-abrv,annot}
	}

\end{document}